# Steering a Predator Robot using a Mixed Frame/Event-Driven Convolutional Neural Network

Diederik Paul Moeys[1], Federico Corradi[1], Emmett Kerr[2], Philip Vance[2],
Gautham Das[2], Daniel Neil[1], Dermot Kerr[2], Tobi Delbrück[1]
[1]Institute of Neuroinformatics, University of Zürich and ETH Zürich, Switzerland
[2]Intelligent Systems Research Centre, University of Ulster, Londonderry, Northern Ireland

*Abstract*—This paper describes the application of a Convolutional Neural Network (CNN) in the context of a predator/prey scenario. The CNN is trained and run on data from a Dynamic and Active Pixel Sensor (DAVIS) mounted on a Summit XL robot (the predator), which follows another one (the prey). The CNN is driven by both conventional image frames and dynamic vision sensor "frames" that consist of a constant number of DAVIS ON and OFF events. The network is thus "data driven" at a sample rate proportional to the scene activity, so the effective sample rate varies from 15 Hz to 240 Hz depending on the robot speeds. The network generates four outputs: steer right, left, center and non-visible. After off-line training on labeled data, the network is imported on the on-board Summit XL robot which runs jAER and receives steering directions in real time. Successful results on closed-loop trials, with accuracies up to 87% or 92% (depending on evaluation criteria) are reported. Although the proposed approach discards the precise DAVIS event timing, it offers the significant advantage of compatibility with conventional deep learning technology without giving up the advantage of data-driven computing.

## I. INTRODUCTION

The DAVIS is a neuromorphic camera which outputs static Active Pixel Sensor (APS) image frames concurrently with dynamic vision sensor (DVS) temporal contrast events [1][2]. DVS address-events (AEs) asynchronously signal changes of brightness. Deep neural network (DNN) technology use has become widespread and there are a large number of tools available for application of this technology. Our aim in this study was to develop the simplest-possible use of DNN technology as applied to DAVIS sensor output. We wanted to preserve something of the data-driven nature of the DAVIS DVS output together with maximum compatibility with existing DNN tools. The aim of this paper, which was a work in parallel with [3] and more functional alternative to [4], is to concretely explore an application in robot navigation.

## II. SETUP AND RECORDINGS

The purpose of the network is to steer the predator robot in the direction of the prey robot. The simplest possible implementation of the predator needs to recognize in which of the 3 vertical regions of the field of view the prey is. These are **L**eft/**R**ight/**C**enter and **N**on-visible (**LCRN**). The *N* output signals that a search for the prey should be initiated. This approach is inspired by the seminal work in [5], [6] and most directly by the forest trail detection and following work of [7], but applied in an indoor scenario with an event-based vision sensor. This limited dimensionality of 4 outputs was determined as the minimal useful case; other networks that also output the prey distance or analog position in the field of view were ruled out by the limited availability of training data at different prey distances. The following two sections introduce the setup used to gather the 500'000 images dataset to train the Convolutional Neural Network (CNN) to perform the task.

### A. Robots

**Fig. 1A** shows the two Robotnik Summit XL mobile robots [8] used in the experiment and the arena. These 750 x 540 x 370 mm robots have four omnidirectional Mecanum wheels that allow skid-row kinematics thanks to small rollers on them. They weigh 40 kg and they can move up to 3 m/s (4 body lengths/second, or the equivalent of 60 km/h for a 4.5 m long car) in the 9.5 x 6.7 m arena. Our experiments were limited to maximum 1.5-2 m/s to prevent possible damaging crashes. The robots are fitted with an Inertial Measurement Unit (IMU) and a laser scanner to detect and avoid collisions. The robots are controlled through the Robot Operating System (ROS) framework [9], running on the embedded PC with Linux with Intel Core i7 processor. They communicate through WiFi 802.11n to allow access to their operating system. The DAVIS sensor is mounted on the predator robot through a simple mounting hole with lock/wing nut. The 2.6 mm wide angle lens provides a horizontal field of view of 81 degree**s.** Lighting was varied by turning the room lights on and off, but the Vicon tracking system, which caused flicker highlights on the floor, was turned off. The windows were not shaded and experiments were conducted under sunny and cloudy conditions. The floor is specular and has stripe patterns, and there are background objects above the walls of the arena. **Fig. 1B** shows the overall system architecture of the predator robot as described in later sections.

### B. Recordings and preprocessing of the data

Twenty DAVIS recordings with a total duration of about 1.25 hour were obtained by driving the two robots in the robot arena of the University of Ulster in Londonderry, as seen in **Fig. 1A**. The predator robot, fitted with the recording DAVIS, followed the prey robot, initially by teleoperation and later autonomously. The prey robot was teleoperated or controlled by a semi-random policy under laser range finder control. DVS and APS data was obtained under conditions to cover variations in lighting, relative position and distance between robots and speed. Various arrangements of background objects such as a black wheelchair and the interference of people walking in front of the camera were recorded.

This research is supported by the European Commission project VISUALISE (FP7-ICT-600954), SeeBetter (FP7-ICT-270324) and Samsung. We would like to thank the Sensors group at INI Zürich, Luca Longinotti from iniLabs GmbH, and the Intelligent Systems Research Centre of the University of Ulster.

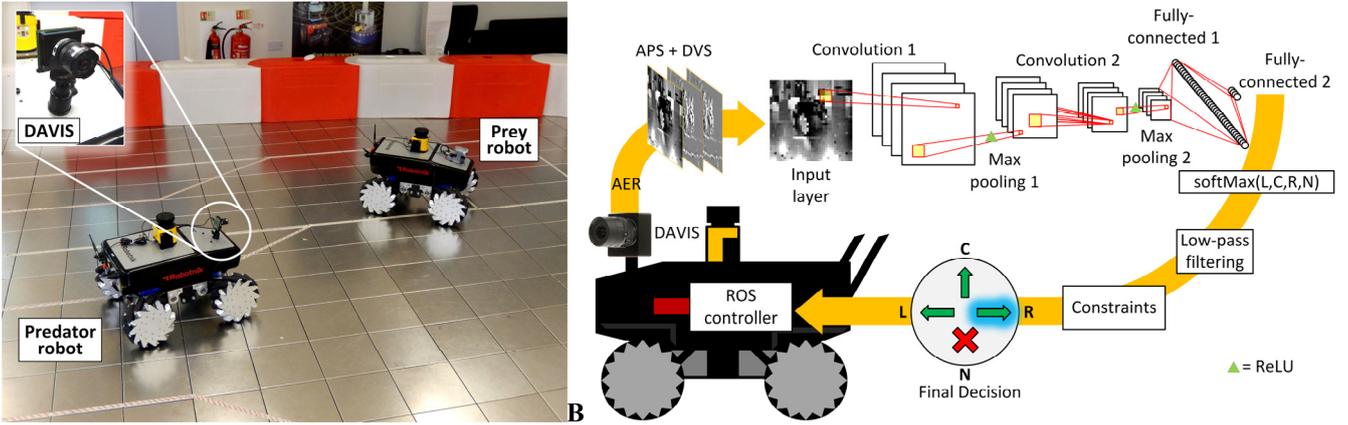

Fig. 1 **A**: Summit XL predator (left) chasing the prey (right) in the robot arena of the University of Ulster, Londonderry. **B**: overall closed-loop system: the DAVIS sensor generates APS and DVS data which is alternately fed to the 4C5-R-2S-4C5-R-2S-40F-R-4F convolutional neural network. The results are filtered and the final decision is used in conjunction with the laser scanner output to control the Summit XL behavior in the ROS controller.

The 240 x 180 APS frames were captured in global shutter-mode using auto-exposure (with typical exposure times of 5 ms) from the sensor at an average of 15 fps. A 36x36 CNN input image size was selected as the minimum size by which the robot can still be recognized by human eye and the easy divisibility of the pixel number in the three steering regions.

DVS data is integrated to 36x36 frames as 2D histograms obtained by integrating 5'000 ON and OFF events in 200 possible gray level values, i.e. starting from a pixel value of 0.5, each DVS ON event increases the gray value by 1/200 and each OFF event decreases it by -1/200. The subsampled pixel address (2D histogram bin) is computed by integer division of the event coordinates. Since the DVS frames are sparse, active DVS frame pixels accumulate about 50 events.

The APS frames are resized down to 36 x 36 with nearest-neighbor interpolation. We used nearest neighbor rather than more accurate methods for computational efficiency targeting embedded application on low power processors. Using jAER [10], the software that processes DAVIS data, both DVS and APS data are converted into uncompressed .AVI video format for later pre-processing with MATLAB.

In total, about 75'000 APS and 275'000 DVS frames were obtained. The data was inflated to 500'000 frames by creating falsely over- and under-exposed APS data by shifting the gray values of the frames by a fixed amount and by clipping the data out of range. This increases the amount of training data and the robustness of the network to new exposures. At the same time, this data augmentation balances out the APS/DVS frames ratio for training, bringing it up to 45% and 55% of the whole training set respectively. An example of the raw recording along with APS and DVS frames is shown in **Fig. 2**.

The ground-truth positions of the prey robot were obtained by manual labelling of the robot position in jAER by capturing the mouse pointer position on the screen during playback of the recordings (using the jAER filter *TargetLabeler*). Depending on which third of the visual field the mouse pointer falls within, the LCRN label is assigned to the frame. 11% of L, 18% of C, 15% of R and 56% of N compose the final training image dataset, which corresponds to the first 80% of each recording.

The remaining 20% of each recording, with LCRN percentages differing by only up to 2%, is used as test set.

Frames are not shuffled randomly before the train and test set separation, because frames are consecutive in time, and so they are highly correlated. Therefore, presenting frames in the test set that are consecutive to those in the training set would result in false higher accuracy, since the network was trained on very similar images, but reduce its robustness to new inputs.

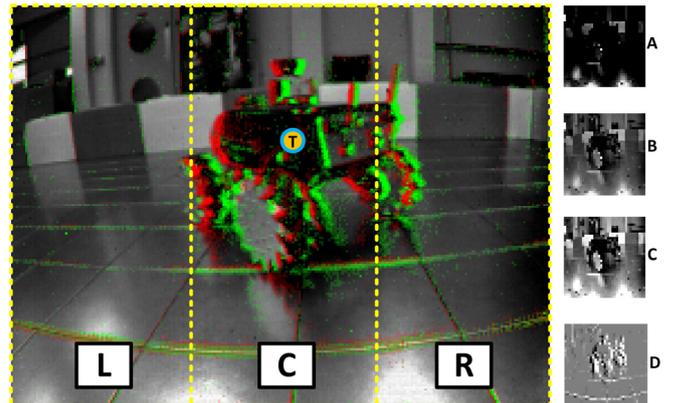

Fig. 2 Example of raw recorded data with overlaying APS gray value data and DVS data (red are OFF events, green are ON events). The field of view is divided into the three regions and the target is labelled. A, B and C are the extracted 36 x 36 APS frames with falsely created exposures. D is a subsampled DVS histogram.

### III. NETWORK TOPOLOGY, INPUT NORMALIZATION, TRAINING, AND ERROR ANALYSIS

We used a CNN because of its proven performance with image recognition [11]. In initial feasibility studies, we first tried to train separate CNNs to process the APS and DVS frames. However, the limited amount of training data invariably resulted in overfitting the training data. To increase the amount of training data, we found that training a single CNN driven sequentially by both types of frames resulted in higher overall accuracy. This approach also simplifies the software architecture. As shown later, it results in CNN first-layer feature detectors that detect robot features in both the APS and DVS frames.

Correct input normalization of DVS and APS frames proved essential for training a network that would work reliably. APS frame input pixel values $F$ are rescaled to the new range $R_n$ 0 to 1 with (1) to obtain the normalized pixel values $F_n$:

$$F_n = R_n \frac{F - \min(F)}{(\max(F) - \min(F))} \qquad (1)$$

Since DVS histograms are not real images, their histogram level which corresponds to zero events (gray in the image) is held at 0.5 (half of $R_n$), to avoid unwanted flickering, dependent on the ratio of ON and OFF events collected. Variation of such gray background would just complicate the recognition task. The DVS histograms are then clipped at three times their standard deviation σ computed around the mean μ of the originally acquired DVS histogram. This clipping removes outliers and allows the DVS histogram to cover the full $R_n$ keeping 99.7% of the information. The entire dataset extraction and generation from 1 hour of recordings through MATLAB takes about 90 minutes. **Fig. 3** shows normalized LCRN examples for APS and DVS.

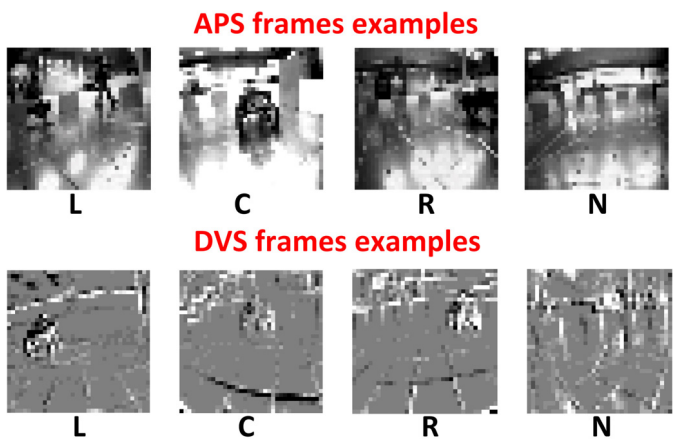

Fig. 3 Normalized LCRN examples for APS and DVS at 36 x 36 resolution.

To overcome the speed limitations of the MATLAB toolbox, previously used in [3], the CNN was trained with the Caffe framework [12]. The CNN consists of the following layers: a 36 x 36 input layer, a convolution layer with $N$ output feature maps with $n$ x $n$ kernels (denoted here as $NCn$), a max pooling layer with stride 2 (denoted as $2S$), another convolution layer with $M$ output feature maps with $m$ x $m$ kernels ($MCm$), another max pooling layer with stride 2, a fully connected layers of $p$ neurons (noted as $pF$) and finally a 4-neuron output layer that is fully connected to the previous layer. The activation function at the end of the convolution and at the output of the fully-connected layers was tested with both sigmoid activation (noted as $S$) and Rectified Linear Unit (ReLU) activation (noted as $R$) types. The network was trained using a softmax loss function on the output layer. The size of the network was chosen through a manual optimization process aimed at the minimum number of features required to perform the task with acceptable accuracy. The performance of some of the various networks explored for the task is shown in **Fig. 4**. Running 100'000 training iterations required about 40 minutes of compute time on the largest CNN in Caffe running in an Ubuntu VM, using CPU-only mode. We used a VM for convenience, since this VM could easily be shared among the authors and moved to different PCs.

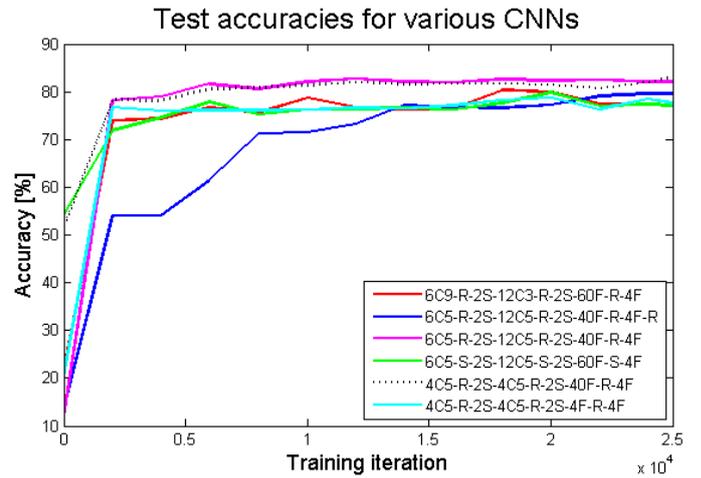

Fig. 4 Testing set accuracies of various networks versus number of training iterations. The selected runtime network is plotted with the dotted line.

The maximum achieved accuracy was 93% on this training dataset and 87% on the test dataset. Inaccuracy was due to a number of reasons, summarized in **Fig. 5** and the rest of this section. The first consists of DVS histograms containing no information: even though no movement takes place in the scene (both robots pause for a moment), the DVS still integrates 5'000 uncorrelated events due to the leakage in the reset switch of the pixel [1]. The result can be seen in **Fig. 5**I. These frames are however relatively few since the reset switch leakage of the pixel causes ON events only every 10 seconds. With 240*180 pixels (43'200 in total), the time to reach 5'000 uncorrelated events is just over 1 second. This noise frame rate is well below the normal DVS frame-rate but still affects the over accuracy on the dataset. These background activity DVS frames could be easily filtered out by the jAER *BackgroundActivityFilter*, but during our experiments we did not use this filter.

Similar to no-movement in the scene, which makes the DVS effectively blind, DVS histograms can be integrated at a higher frequency if both the predator and the prey are still but another object moves, creating events. If a person passes by in this situation, for example, the prey is falsely predicted to be invisible (**Fig. 5**J). Other noise events can disrupt the proper computation on DVS histograms. The main one is parasitic capacitive coupling and non-optimal biasing. This coupling between the global electronic shutter and the DVS event generation mechanism causes a burst of DVS events on each frame [2] and creates events correlated with the sample rate of the APS, filling up the 5'000 events allowed in the DVS histogram, sometimes covering up the prey robot (especially if far away). This noise can be recognized by the scanning lines (**Fig. 5**O,P) which indicates strong coupling activity which saturates the event arbiter. In some images, when the robot is far away, this artifact overlaps the prey (**Fig. 5**O,P) and sometimes the prey is too small even for human detection (**Fig. 5**C,D,K). In other cases features of the background can take over and be recognized as the prey (**Fig. 5**G,H,L). An interesting example is a black wheelchair that was present in the corner of the robot arena. This looked in the downsampled image like the prey robot and the only way to recognize it was with correlation with previous frames where the prey position is known to be different. Since the network learns without any frame-to-frame correlation, then this is a problem that remains unsolved for frames where the robot is far away and of very similar shape to the black wheelchair.

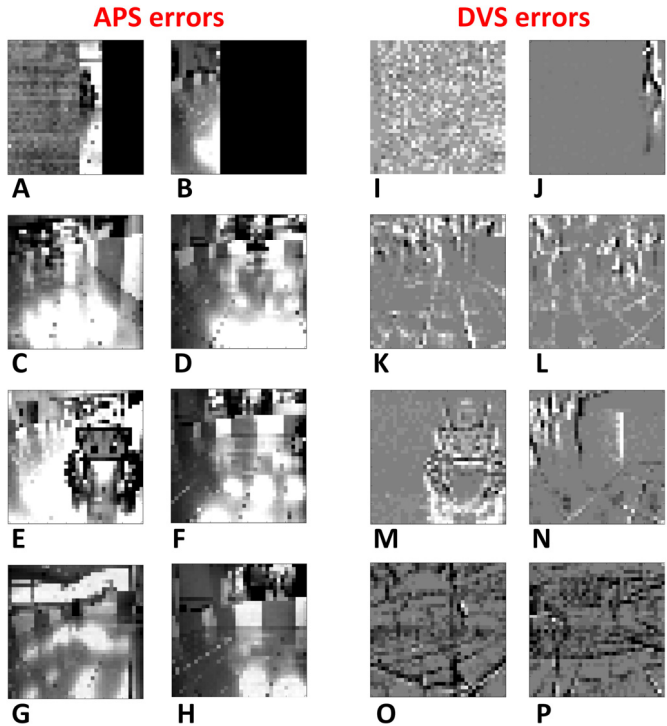

Fig. 5 Examples (36 x 36 images) that cause wrong network predictions. A and B: corrupted and dropped APS frames. C and D: prey far away and overlapping with other similar shapes. E and F represent ambiguous APS frames: in E, they prey is very close, covering two thirds of the field of view, in F, only one wheel of the prey is present. G and H: false positives caused by a black wheelchair and a high-contrast poster. I: DVS histogram which integrated random background noise due to no movement present in the scene. J: DVS histogram where the prey and predator are still but a moving person triggered the DVS frame integration. This makes the prey effectively non-visible from the DVS output. K: prey too far away to have good resolution. L: wheelchair appearing as a robot to the network. M and N: same ambiguity of E and F. O and P: noise activity integrated in the DVS histogram due to coupling between APS and DVS in the DAVIS sensor.

Furthermore another factor which deteriorates APS images is the occasional corruption and partial loss of APS frames (**Fig. 5**A,B). This problem in our experiments was due to incorrect setting of the USB buffer sizes. Since the host side buffers were set too small and APS data arrives in large bursts, some data is occasionally dropped, including frame start events, resulting in corrupted frames. When a black stripe covers the prey, it is impossible even for a human, without previous frame correlation, to know the current position of the robot. In our experiments this problem was handled by low-pass filtering the network predictions.

Finally, the main reason for errors consists in the ambiguity of the frame when the robot crosses boundary regions and the label oscillates around it. Other ambiguities examples are when only a wheel of the robot is visible at the extreme edge of the image or when the prey is close enough to cover two thirds of the field of view of the predator (**Fig. 5**E,F,M,N). Since the data is hand-labelled using a subjective interpretation of the robot position in the image, the ambiguous frames are bound to deteriorate the overall accuracy. To prove the importance of this second factor, the accuracy of the test-set was recomputed to eliminate the most ambiguous frames in which the labelled target position is within 1 pixel (in the 36 x 36 image, corresponding to about 6 pixels in the 240 x 180 original image) of the four boundary region. This on average improves the accuracy by 3%. Increasing this error margin further improves the accuracy of the networks. The rest of the ambiguous images are the ones where the prey robot is very close to the predator and more than one LCRN region is covered by it. There are however still some frames for which the correct decisions are very obvious to human eye where the network fails for no apparent reason.

Test and train accuracies are close to each other (on average there is a 5% difference between the two) and indicate that overfitting is minimal. Overfitting was also minimized by using dropout (randomly setting to zero in each iteration a fraction of the weights to reduce weights' co-adaptation) of 20-30% of the first fully-connected layer. The gradient-based optimization method chosen was "Adam" [13], to deal with the large amount of training data.

The size of the network was chosen observing the effect of each of its parts. Regarding the convolution, increasing the size of the square kernel increases the network's accuracy. However, computing time and overfitting eventually increase too. A size of 5 x 5 was found to be optimal. The number of feature maps per convolution layer was reduced to the minimum necessary. We found that a surprisingly small number of features were needed. When we used more than about 4 features per layer, kernels started to repeat or ended up with near-zero weights. The optimum number of neurons in the first fully-connected layer was 40; providing more units did not increase accuracy but leaving out the first fully connected layer significantly reduced accuracy. A ReLU activation function provided similar accuracy as sigmoidal activation, but was much less likely to become stuck in a local minimum where the network would detect only the class most present in the training data. A lower accuracy is also observed for networks terminating with ReLU units, therefore this activation function was used at convolution outputs and the first fully-connected layer output. Since the network decision is always taken as the maximum activated output unit, this finding does not make sense at first, but was probably the result of back-propagating the unbounded ReLU outputs or not backpropagating error reduction from negatively-activated ReLU's.

After exploring many different combinations of parameters, we finally settled on a runtime $4C5$-$R$-$2S$-$4C5$-$R$-$2S$-$40F$-$R$-$4F$ architecture providing low computational cost and acceptable accuracy. Its kernels of the first and second convolution layer are shown in **Fig. 6.** The kernels in the first convolution layer extract the most basic features of this predator/prey context with spatially-confined filters. As a matter of fact, kernel 0 and 1 seem to highlight the high contrast of the wheels of the prey and kernel 2 and 3 the edges of the arena walls above and to the right of the robot. The kernels of the second convolution layer combine the first layer features, but their interpretation is more difficult.

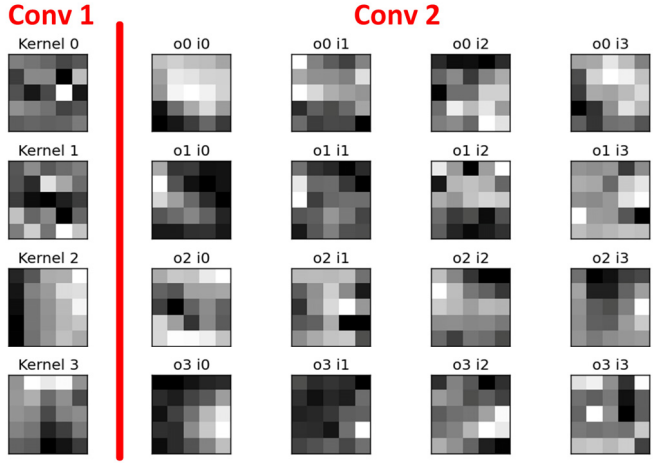

Fig. 6 Left of red line: 4 kernels of the first convolution layer 4C5 for the chosen network (4C5-R-2S-4C5-R-2S-40F-R-4F). Right: 16 kernels of the second convolution layer 4C5. For example, o2 i1 is the kernel for output unit 2 of the second convolution layer that filters output feature 1 of the first subsampling layer.

In **Fig. 7**, the guided-backpropagation method [14] of saliency visualization was employed to determine the parts of a particular input image that resulted in a strongly winning C activation. The guided backpropagation process, which is a mixture of backpropagation based on the input data and a deconvolution of the gradient, hides the influence of negative gradients which decrease the activation of the target neuron while highlighting regions that strongly affect the target neuron. The example shows that for this input, the robot wheel and dark body cause the strong activation, while other features such as the grid and highlights on the floor and the walls and background are ignored.

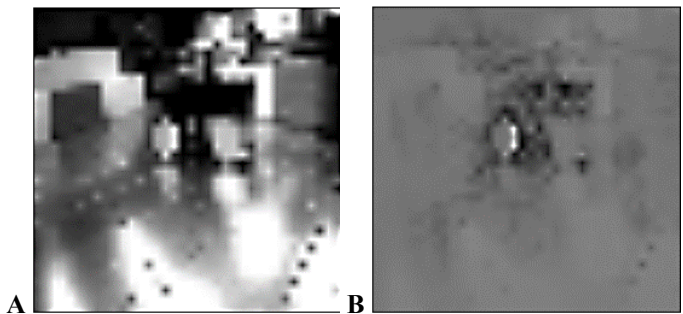

Fig. 7 Saliency visualization using the guided-backpropagation method. **A:** the input image (36 x 36) presented to the network**. B:** Saliency overlay, with highly salient regions transparent to expose the input data that caused the active class label (robot in center).

## IV. JAER IMPLEMENTATION

jAER is a software project for visualizing, recording, and soft real time processing of event based sensors [10]. A CNN runtime engine along with necessary utilities such as the target labeler were implemented in jAER as the Java package *eu.visualize.ini.convnet*. The pre-trained weights of the kernels and of the fully-connected layer are loaded from an XML file describing the network into jAER. A python script *cnn_to_xml.py* in the jAER *scripts/python/caffe_utils* folder reads the Caffe network architecture and weight files to produce the XML file. The CNN takes as input the downsampled and normalized 36 x 36 APS frames and the subsampled and normalized 36 x 36 DVS histograms of 5'000 events. The forward CNN pass requires between 1-2 ms compute time for the chosen network size on the robot PC, which has a Core i7 processor running Java JDK1.8. The final decisions of the CNN, along with every intermediate activation map, can be visualized during runtime as shown in **Fig. 9.** When playing back labeled data in jAER, the error rates for both APS and DVS frames can be computed online and compared with the offline results of the training.

## V. ROBOT COMMUNICATION

jAER runs concurrently with ROS on the robot. jAER sends its steering decisions to ROS using User Datagram Protocol (UDP). The two integers are sent as two bytes. The first is a sequence number 0-255 to check for lost datagrams and the second is the robot steering direction encoded as L=0, C=1, R=2, N=3. Communicating through UDP allows jAER to run independently as a server on the same computer and not as a ROS slave node. This way its processing can be monitored and controlled without modification of jAER's architecture. The UDP messages are sent at each novel decision. The computations run in jAER at a jAER processing rate of 240 Hz, which is near the fastest that DVS frames are generated. Thus, even though jAER processes the DAVIS data in packets, most packets result in no decision and few packets produce more than one decision. Processing at a lower rate in jAER, with larger packets, would result in multiple decisions per packet, all sent in a very short time, which would not make sense. **Fig. 8** shows the measured interval distribution for commands sent to ROS for the trial run #7. The median communicated decision rate was 91 Hz. The APS frame rate was about 15 Hz, and therefore the other 76 Hz rate of decisions was caused by DVS frames. With a DVS frame size of 5'000 events, an average rate of DVS events of about 76 Hz*5'000 = 380'000 DVS events per second (**eps**) rate is consistent with our observations of the average DVS event rate during this recording. For the fastest trial run describe later, where the robot ran at 2 m/s, the average DVS event rate was about 600 keps, resulting in average DVS frame rate of about 120 Hz, although peak rates often reached the jAER processing cycle rate of 240 Hz.

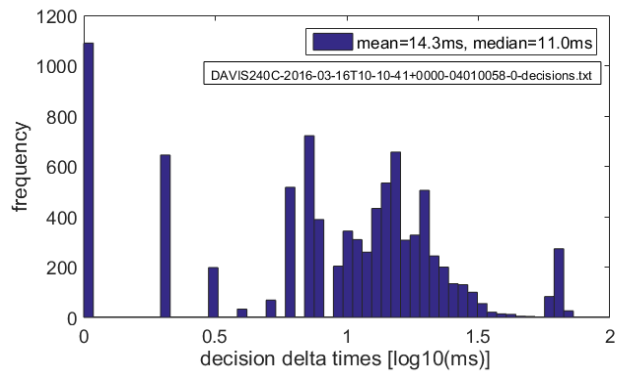

Fig. 8 Decision output interval distribution.

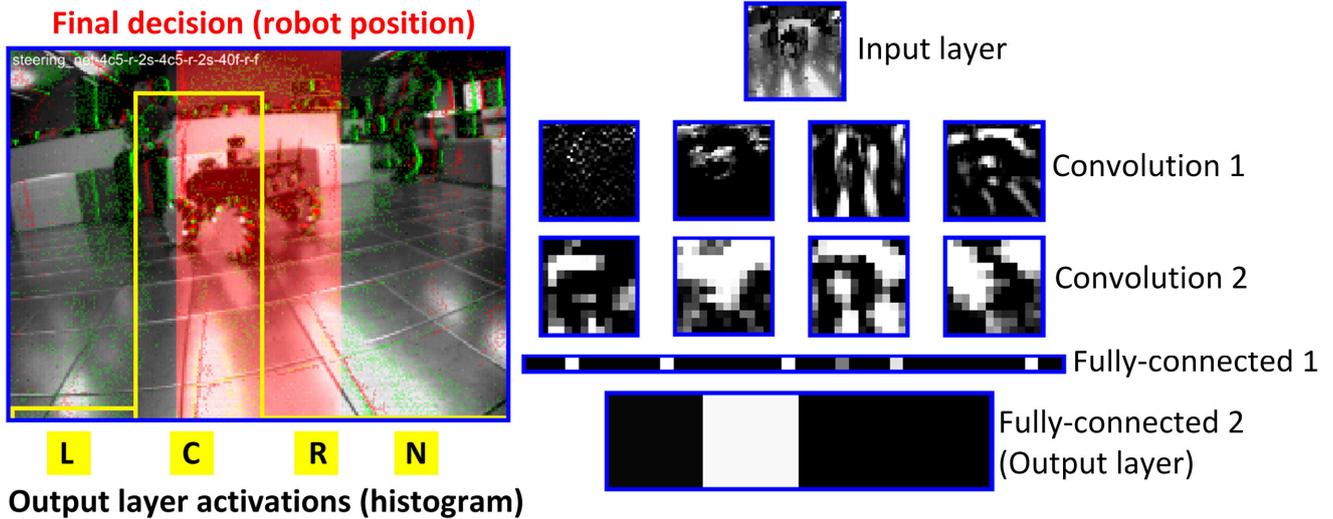

Fig. 9 JAER visual output of the jAER CNN's processing (4C5-R-2S-4C5-R-2S-40F-R-4F). On the right-hand-side: the activations maps of input, convolution and fully-connected layers respectively. On the left-hand-side: final network decision: the four analog outputs of the four-neuron of the second fully-connected output layer are represented as an histogram of four bars in yellow. The red rectangle shows the final result of the softmax operation highlighting the section of the field of view where the prey is in. In case of non-visibility, no red rectangle is displayed. See video [12].

## VI. BEHAVIOR AND DECISION FILTERING

The behavior of the predator robot allows several heuristic filtering options to reduce noise in the CNN decision output. Low-pass filtering is possible and an advantage of real-time, since frames are consecutive and strongly correlated.

The low pass filtering maintains an analog LCRN state for each possible LCRN CNN decision output. The output (winner) of the low pass is the LCRN state which has the largest value. These LCRN states are bounded from 0 to 1. A parameter $\alpha$ specifies a step size. For each decision, the corresponding state is increased by $\alpha$ and all other states are decreased by $\alpha$. For our trial runs when low-pass filtering was used, we set $\alpha=0.25$. For example, if the current winner was L with state value 1 and state C was at 0.25, it would require 2 consecutive C decisions for C to be the new winner, because L would follow the sequence 0.75, 0.5, and C would follow the sequence 0.5, 0.75. Thus, after 2 decisions, C would become the new winner.

When the ROS controller receives L or R decisions it unconditionally turns left or right respectively at an angular velocity of $\pi/3$ rad/s and with the maximum allowed linear velocity (chase mode). The maximum linear velocity suitable for the arena used in these experiments is about 1.5 m/s. Trials at 2 m/s showed that safety distances in the presence of an obstacle need to be increased to avoid crashes, effectively reducing the useful area of the arena by more than a meter per side. At higher speeds it is also difficult for the teleoperated prey to evade the predator. Finally, the linear speed of the predator robot is also regulated by an underlying model of potential fields, an obstacle-avoidance algorithm for path planning. According to this algorithm a vector field is established over the area visible by the laser scanner and obstacles have a repulsive force dependent on distance that reduce the linear velocity of the robot. The aim of the robot behavior is forcing the prey to be in its center C (which corresponds to the central 27° of the FOV of the predator with the chosen lens) and accelerating towards it until the minimum safety distance is detected by the laser scanner in the center 40° FOV. Upon initialization, if the predator does not see the prey, or if it loses it for more than 5 seconds, the predator goes into search behavior (wander mode) and moves around randomly. If instead the prey becomes non-visible after being on its left or right it spins in the direction it last saw the prey (either left or right respectively) at an angular speed of 1.5 rad/s and zero linear speed (rotate mode). This behavior motivates the first constraint to the CNN output: CNN decisions that indicate that the prey is again visible on the opposite side of the field of view from which it disappeared can be discarded. Two more logical constraints can be applied. The first one is that the prey cannot switch instantly from center to non-visible. The final constraint is that the prey cannot pass from left to right without passing the center. This logical information can be used as post-processing of the CNN's output to increase its accuracy, at least in the autonomous scenario (some the data in the training set were generated by teleoperation of the predator).

The robot proceeds forward to the detected prey until the laser scanner detects imminent collision and stops. If the robot dashed forwards while detecting the prey in its center and a collision was detected, the prey is considered captured (prey caught mode). After this, the predator robot spins in the correct direction to center the prey and the prey moves away, controlled by the user or by the automatic navigation protocol. The chase starts again after a few seconds. All information about the state of the robot behavior (chase, wander, rotate and prey caught modes) is sent back to jAER through UDP, for display and recording.

## VII. CLOSED LOOP RESULTS

Once a CNN network with acceptable accuracy was trained, the robot control loop was closed by letting the predator robot run following the computed steering commands. While the prey was manually driven by an operator at the University of Ulster, the predator was being driven by decision outputs. Initially we controlled and monitored the jAER CNN interface live from Zürich through a TeamViewer connection. The last 8 trial runs were conducted on site.

A video of one chase sequence (trial run #8) is available at [15], and a video of the CNN activity during part of this sequence is available at [16]. Video [15] shows the robot arena in which the predator drives at 1.5 m/s when moving forwards. Low-pass filtering with α=0.4 and heuristic decision filtering, as described in the previous section, were used to smooth the decision output. The synchronized jAER running the CNN with both APS and DVS frames is also shown. The ground truth prey locations were labeled and online accuracy statistics are shown in the video. These statistics are based on the unfiltered raw CNN decision output, before low pass and heuristic filtering. Video [16] shows a view like **Fig. 9** of the network activity during part of the same run. The topographic arrangement and responses of the convolutional layers becomes more obvious in this video.

For each of the eight trial runs, the ground truth was labelled offline. Speed was varied from 0.5 to 2 m/s and decision constraints were not applied in two trial runs. In three trial runs the lighting conditions were also altered (certain lights were turned off) to check for robustness, although no difference was noticed due to the auto exposure control of the APS frame capture of the DAVIS and the automatic local gain control of the DVS pixels. The decisions of the network were recorded and timestamped, to estimate the accuracy, which is shown in **Fig. 10** for the first seven trial runs. Accuracy is plotted against the number of pixels $p$ in the boundary overlap, that is to say, the number of pixels that constitute the margin within which the decision is can still be considered correct if it is either one of the two neighboring LCRN regions forming the particular overlap and the ground truth falls within this same margin. If, for example, the ground truth labeled location is outside the C region but to the right or left within a distance $p$ of the edge, then a C decision is still considered correct. And similarly, if the ground truth location is within p of the outer edges (indicating that the robot is only partially visible), then an N decision is still considered correct. Any L, C, or R decision when the ground truth label is N is a false positive and is always considered incorrect. These criteria are illustrated in **Fig. 11**.

It can be observed that increasing the overlap regions where the target location is ambiguous increases the accuracy. The various accuracies start from different levels since the number of ambiguous frames changes in every run. If the prey robot moves slower, then it is more likely to be mostly covering more than one third of the field of view generating ambiguous frames. Or it could be around the central region and oscillating in position around its boundaries. These cases quantitatively lead to a higher error rate as compared to the hand-labelled ground truth but qualitatively they are irrelevant as the robot will barely move if the decisions oscillate around a boundary. From **Fig. 10**, it can be seen that the use of constraints improves accuracy by about 6-7%. Interestingly, it can also be noted that when the maximum speed of the robot is increased, the accuracy is also increased. This is probably due to the fact that since the predator is faster, the prey has less time to escape. Therefore, the prey cannot move out from the predator's center C into the ambiguous steering regions that decrease accuracy. The achieved accuracies are comparable to [7], although the context of the robot arena and the application (the chase scenario) are probably of lower complexity than natural forest trails.

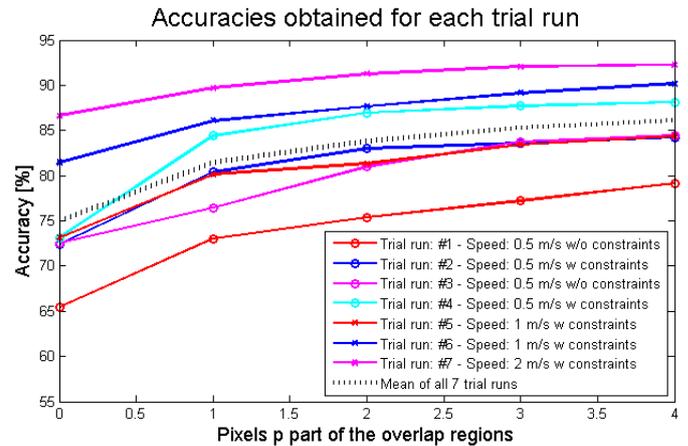

Fig. 10 Accuracies of the seven trial runs versus number of pixels $p$ part of the overlap regions of the 36 x 36 image.

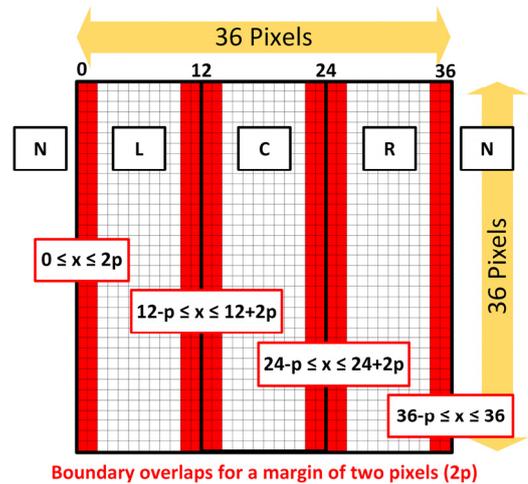

Fig. 11 Example 36 x 36 frame and overlap regions (shown in red) of width multiple of $p$, within which the decision is still considered correct if the ground truth labeled target location is within either one of the two neighbouring LCRN regions forming the particular overlap (and the ground truth labelling falls within it).

The DVS frame error rate was slightly higher than that of the APS frames. For example, in trial run 8 (not included in **Fig. 10**), the overall raw error rate before decision filtering was about 13%; for APS the error rate was about 9% and for DVS the error rate was about 14%. If we decreased the DVS frame to only 1'000 events, then the DVS error rate rose to 21% but the DVS frame rate rose to about 500 Hz. Thus there is a tradeoff between DVS frame size (in events) and decision accuracy. This tradeoff is to be expected, because integrating fewer DVS events results in a higher sample rate but a more quantized image, also with statistics on which the CNN was not trained.

## VIII. CONCLUSION

This paper proposed a method for combining the well-developed frame-based field of convolutional neural architectures with the data-driven processing of neuromorphic engineering using a simple approach aimed to be maximally compatible with existing training tools. With a processing rate proportional to the scene activity, encoded in the number of events, it is possible to reduce the amount of computing power whenever it is not needed. Although using DVS frames throws away the precise temporal information contained in the events, it still provides the advantages of the local temporal contrast response of the DVS pixels, which extracts features robustly in wide dynamic range lighting conditions. These conditions are a problem for conventional APS frames, e.g. as observed in forest scenes in [7].

In contrast to the training based on a simulated road in [6], learning user steering commands in [5], and learning trail versus non-trail using three cameras mounted on a human's head while they walk along trails in [7], our training was based on hand labeled prey target locations. This was feasible because there was only one target and we had only to label its 2D location. This labeling could be done almost in real time during playback, so as a fraction of the entire training process it represented only a tiny fraction of the effort.

The developed chase system ran robustly even with accuracies of around 80%. Our main finding and biggest surprise was the small size of CNN required for solving this problem. More surprisingly, it reliably detects the absence of the prey with about the same accuracy as it detects the presence and location of the prey. That means that this tiny CNN must detect the absence of a conjunction of features characterizing the prey in a rather complex (but static) background scene, which is a much more difficult task. The selected runtime network has only about 10'000 parameters. If we define each necessary multiply or add as one operation, then the forward pass requires only about 350'000 operations. (The current jAER CNN implementation has a much higher operation count because of matrix indexing computations.) This low operation count puts the computational cost well-within the range that could be serviced by small embedded application processors in a more optimized implementation. Although we did not explore it in our experiments, the computational cost could have been substantially reduced by adaptively controlling the DAVIS APS frame capture. For example, APS frames might be triggered only when the DVS event rate is low, or when the DVS output layer produces an ambiguous analog decision.

In future work, the size of the prey robot could also be taken into account to determine the distance from the predator robot. The only reason this was not done in the current project was lack of time and initial training data. The apparent height of the prey robot in the recording could also be used to infer the distance of the prey, however, this was not possible with the available recordings. This height constancy is due to the camera mounting, which points the recorded scene's vanishing point to the horizon line, so that the position of the prey does not vary significantly in height with prey distance. The more general problem of identifying the walls of the arena and the robot's direction and distance from them could also be studied. These specific problems for the robot arena could be considered as prototypes for vehicle and path detection in autonomous driving (i.e. on roadways or factories), but with the advantage of smaller datasets and networks that are much easier to study.

The runtime jAER CNN implementation and Caffe conversion script are open-source. The database of recordings, trial runs and Caffe datasets are available on request.